\documentclass{IOS-Book-Article}

\usepackage{amsthm,amsfonts}
\usepackage{mathrsfs}
\usepackage[OT2,OT1]{fontenc}
\makeatletter
\@ifundefined{l@nohyphenation}
{\chardef\l@nohyphenation\@cclv}{}
\newcommand{\cyrillic}{%
\language=\l@nohyphenation
\fontencoding{OT2}%
\fontfamily{wncyr}%
\selectfont}
\makeatother
\DeclareTextFontCommand
{\textcyrillic}{\cyrillic}

\newtheorem{mydef}{Definition}

\begin{document}

\pagestyle{headings}
\def\thepage{}

\begin{frontmatter}       

%
\title{A General Framework for Describing Creative Agents}

\author[A]{\fnms{Valerio} \snm{Velardo} \thanks{Corresponding Author: Valerio Velardo, University of Huddersfield,
Huddersfield, UK; E-mail:
valerio.velardo@hud.ac.uk.}} and 
\author[A]{\fnms{Mauro} \snm{Vallati}}

\runningauthor{Velardio and Vallati}
\address[A]{University of Huddersfield, UK}

\begin{abstract}
Computational creativity is a subfield of AI focused on developing and studying creative systems. Few academic studies analysing the behaviour of creative agents from a theoretical viewpoint have been proposed. The proposed frameworks are vague and hard to exploit; moreover, such works are focused on a notion of creativity tailored for humans.

In this paper we introduce General Creativity, which extends that traditional notion. 
General Creativity provides the basis for a formalised theoretical framework, that allows one to univocally describe any creative agent, and their behaviour within societies of creative systems. Given   the   growing   number   of   AI   creative   systems   developed   over   recent   years,   it   is   of  
fundamental   importance   to   understand   how   they   could   influence   each   other   as   well   as  
how   to   gauge   their   impact   on   human   society.   In   particular,   in   this   paper   we exploit the proposed framework for (i) identifying different forms of creativity; (ii) describing some typical creative agents behaviour, and (iii) analysing the   dynamics   of  societies   in   which   both   human   and   non-human creative   systems   coexist.
\end{abstract}


\end{frontmatter}
\pagestyle{empty}

\section{Introduction}

Creativity is a broad phenomenon which has long been studied by researchers from several different disciplines, such as psychology, philosophy, computer science, and humanities. An agreed definition of creativity is still missing because of the numerous, sometimes contrasting, definitions proposed by experts \cite{meusburger}. However, scholars identified the generation of novel and valuable items as a fundamental property of creative behaviour \cite{mumford}. Up until the 1950s, the focus of research in creativity has been humans only. In the last decades,  
given the increasing importance of computational systems within human society, Artificial Intelligence (AI) systems have been developed that show creative behaviour. This field of AI is called {\it Computational Creativity}.
The aims of computational creativity is to design AI creative agents, and to provide theoretical frameworks necessary to describe and evaluate creative systems. However, while a large number of systems have been developed for generating various creative artefacts, few studies are focused on theoretical tools (e.g., \cite{wiggins2006preliminary,colton4}). Moreover, they are limited to a notion of {\it anthropocentric} creativity, tailored for human evaluation capabilities, and rarely allow to evaluate the interactions within societies of creative agents.


In this paper we introduce the concept of General Creativity, which goes beyond anthropocentric creativity. We use this notion as a basis to build a formalised theoretical framework. Such a framework provides the necessary tools for: (i) univocally describing any creative agent; (ii) studying societies of creative agents, and (iii) characterising different forms of creativity. 
Given the growing number of AI creative systems developed over recent years, it is of fundamental importance to provide a clear and formal description of their components and behaviours. Moreover, it is pivotal to understand how they could influence each other as well as their impact on society, in order to improve our knowledge on the creativity process itself. Therefore, a framework capable of modelling this phenomenon will be essential in the near future.


\section{Background}
Even though creativity is a loose concept, scholars agree that it involves the generation of novel and valuable artefacts \cite{mumford}. Apart from these two elements, researchers usually differ greatly in the way they define creativity \cite{meusburger}. 
Wallas proposed a pioneering model \cite{wallace}, in which the generation of an artefact happens in four stages: {\it preparation}, {\it incubation}, {\it illumination} and {\it verification}. Koestler introduced the concept of {\it bisociation} \cite{koestler}, which explains creativity as the combination of two unrelated initial concepts, in order to generate a new creative product. 
Even if useful, the theory of bisociation does not describe the creative process in a completely structured manner. Finke, on the other hand, addressed the issue by proposing {\it Geneplore} \cite{finke}, 
a model of creativity based on generation and refinement of creative products. 

Boden tackles the topic from a theoretical perspective, offering an insight into different categories of creativity \cite{boden2004}. {\it Exploratory creativity} consists of exploring a given conceptual space, whereas {\it transformational creativity} consists of changing the conceptual space itself. Furthermore, if the artefacts generated are new only to the system, then the agent shows {\it P-creativity}; on the other hand, if they are new to society, the agent is {\it H-creative}. Although these concepts are extremely valuable for the study of creativity, they lack a precise formalisation. Such gap has been filled by Wiggins, who proposes a general framework for creative systems \cite{wiggins2006preliminary}, as well as a search-driven approach to the generation of artefacts, which exploits Boden's ideas \cite{wiggo2}. 

Creativity is not only a human prerogative, but it can also be shown by machines \cite{boden2}. In that regard, Colton and Wiggins define computational creativity as the science of developing computational systems that exhibit behaviours that would be deemed as creative by unbiased observers \cite{colton1}. Computational systems have proved to be effective in the generation of creative artefacts in several domains, such as music  \cite{diaz,chuan,velardo}, visual arts \cite{colton2,cohen} and story/poetry generation \cite{gervas,riedl,colton3}. The synergy between humans and machines obtained by developing computer-aided systems can lead to new and unexpected forms of creativity as well \cite{morris,li}. 

The shift from the view of creativity as dominated solely by humans to a form of creativity in which non-human systems can act as creative agents is slowly happening. However, issues with computational creative systems still remain. The evaluation of artefacts 
is a major problem. Although several valuable works have been published \cite{pease,colton5,colton4,jordanous}, they look at evaluation from the human viewpoint. In the following, we extend the notion of evaluation to non-human systems, by considering a general form of creativity.   






\section{Defining Creative Systems}

Before providing a formal framework for the description of a {\it general} -- i.e., non-anthropocentric -- creative system, we provide a definition of the expected output of any creative system, which is an {\it artefact}.

\begin{mydef}\label{artefact}

An artefact is any element either conceptual or physical, which can be generated.

\end{mydef}

According to Definition \ref{artefact}, we emphasise that the product of a creative system can be either physical (e.g., paintings, sculptures) or conceptual (e.g., scientific theories, ideas). 

Having now given a definition of artefact, we can informally describe a creative system as a system that can create artefacts, and that can be mainly characterised by a process of {\it generation} and a process of {\it evaluation} of such artefacts. Formally, we define a general creative system as follows.


\begin{mydef}
A creative system is a 7-tuple $\mathcal{S} := \langle \mathscr{A}, \mathcal{M}, \mathcal{I}_s, \mathcal{E}_s, \lfloor \cdot \rceil_\mathit{g}, \lfloor \cdot \rceil_\mathit{v}, \lfloor \cdot \rceil_\mathit{u}\rangle $, where:
\begin{itemize}
\item $\mathscr{A}$ is the Artefacts space;
\item $\mathcal{M}$ is the Memory set;
\item $\mathcal{I}_s$ is a specific configuration of Internal constraints;
\item $\mathcal{E}_s$ is a specific configuration of External constraints;
\item $\lfloor \cdot \rceil_\mathit{g}$ is the Generation operator;
\item $\lfloor \cdot \rceil_\mathit{v}$ is the eValuation operator;
\item $\lfloor \cdot \rceil_\mathit{u}$ is the Update operator.
\end{itemize}
\end{mydef}

This definition extends the preliminary framework description given by Wiggins (\cite{wiggins2006preliminary}), in the sense that we consider creative systems which show anthropocentric and non-anthropocentric creativity.

The generation operator considers the constraints, both internal and external to the system, in order to generate an artefact. For assessing the quality of the artefact, and deciding whether it should be provided as output of the system or not, an evaluation operator is used. The evaluation operator also relies on memory, which allows the judgement of {\it novelty} of the artefact, guaranteeing P-creativity, as defined by Boden \cite{boden2004}. Finally, in the framework we include an update process. Although it is not strictly necessary, the update operator, by modifying generation and evaluation processes, can lead  a creative system to show transformational creativity \cite{boden2004,wiggins2006preliminary}.

In the following, we provide a detailed description of each component of a creative system.

\subsection{Artefacts Space}

Our formalisation of the artefacts space $\mathscr{A}$ follows the definition of Universe provided by Wiggins (\cite{wiggins2006preliminary}). 

\begin{mydef}
The artefacts space $\mathscr{A}$ is a multidimensional space, whose dimensions are capable of representing any possible artefact; all possible distinct artefacts correspond to distinct points in $\mathscr{A}$.
\end{mydef}

Contrary to Wiggins' definition of Universe, which represents concepts, we opted for a space populated by artefacts, so that we can account for both physical and conceptual elements. 
Also, an artefact is a ``complete'' concept, in that it symbolises the final product of a creative system. 
Our framework preserves the axioms of {\it Universality} and {\it Non-identity}, introduced by Wiggins. The former indicates that all possible artefacts, including the empty one $\top$, are represented in $\mathscr{A}$. In our interpretation the empty artefact $\top$ can also be the result of a worthy creative process. 
The latter axiom claims that for each possible couple of distinct artefacts $(a_i,a_j)$ contained in $\mathscr{A}$, $a_i$ and $a_j$ are mutually non-identical.



\subsection{Memory Set}

In a creative system, memory $\mathcal{M}$ can be defined as follows.

$\mathcal{M} := \{ x$ $|$ $x \in \mathscr{A}, x$ ${\verb observed } \}$

\noindent
The formalisation represents the set of all the artefacts that the system has observed and is aware of. A system has observed all the artefacts that it already generated, but also artefacts that other creative systems have generated and have shown to it. It is worth noting that a creative system can observe and store, only information of artefacts that it is able to recognise as artefacts. For instance, a human can not hear a musical piece exploiting only ultrasound, thus will not be able to store the corresponding musical piece in memory.
Memory, as we define it, is a fundamental part of a creative system, since it allows to assess the novelty of an artefact with regards to the agent itself. Novelty has been demonstrated to be critical for creativity \cite{sterberg}. Memory is then exploited by the evaluation operator, in order to correctly assess the quality of artefacts.


\subsection{Internal Constraints}
Following the notion of style by Meyer, each creative system generates an artefact choosing patterns from within a set of constraints \cite{meyer}. Some constraints are deeply embedded within the system and cannot be overcome. For example, perceptual/cognitive constraints are biologically rooted within humans, and do not usually change significantly over the lifetime of a person. 
Of course, there seems to be some counterexamples, such as people who experienced sight loss. However, in those extreme cases, the person changes so radically, that she can be seen as a new creative agent. 
The same is true for a computational system, which is constrained by a certain amount of processing power. 
It is possible to replace some parts of the system in order to improve its computing power but this results in a new creative system. 
We define these deeply embedded constraints, which cannot be easily changed in the lifetime of a system, as ``internal constraints".

Internal constraints are formally defined as a set of rules.

$\mathcal{I} := \{ x$ $|$ $x$ ${\verb is }$ ${\verb an }$ ${\verb internal }$ ${\verb constraint } \}$

\noindent
From the general set, it is possible to derive a specific configuration which completely describes the internal constraints of a given creative system; by providing a set of pairs, in which each rule is coupled with a weight, that specifies the relevance of a constraint to the system. 

$\mathcal{I}_s := \{ (w; i)$ $|$ $i \in \mathcal{I}, w \in \mathbb{R}_{[0,1]} \}$

\subsection{External Constraints}
While some constraints cannot be easily changed over the lifetime of a system, others can. We define these as external constraints. External constraints strongly shape both the generation and evaluation processes of creative systems, but they can be modified, thanks to the exchange of information of a system with its peers. For example, culture effectively shapes human constraints. Even though some of these are quite difficult to modify, it is possible to vary them, by having people exposed to a different culture for a long enough period of time. People who begin studying an instrument dramatically modify the way in which they perceive music \cite{gaser}. The case for computers is similar. They are programmable, meaning that they can change procedures and values of their software, according to external pressures.

Just like internal constraints, external constraints can be defined as a set of rules.

$\mathcal{E} := \{ x$ $|$ $x$ ${\verb is }$ ${\verb an }$ ${\verb external }$ ${\verb constraint } \}$

\noindent
Specific configurations of external constraints are represented by a set of pairs; each rule is complemented with a weight.

$\mathcal{E}_s := \{ (w; e)$ $|$ $e \in \mathcal{E}, w \in \mathbb{R}_{[0,1]} \}$

\noindent
$\mathcal{E}^T$ indicates all possible configurations of external constraints, which a creative system can take. $\mathcal{E}^T$ is necessarily a subset of all possible configurations, since internal constraints indirectly delimit the domain of external constraints of a system. For example, as a result of internal constraints, human painters are not able to generate micron-sized painting. Therefore, it is not possible for humans to develop external constraints related to micron-sized painting, since it exceed human capabilities. 

\subsection{Generation Operator}

The generation operator $\lfloor \cdot \rceil_\mathit{g}$ is the part of the creative system responsible for creating new artefacts. It receives both the internal and external constraints as input, and returns an artefact $a$ from the artefacts space $ \mathscr{A} $. Formally, we can describe the generation operator as follows.

$ \lfloor \mathcal{I}_s,\mathcal{E}_s \rceil_\mathit{g}  = a$ $|$ $a \in \mathscr{A} $

$ \lfloor \cdot \rceil_\mathit{g} : \mathbb{R} \times \mathcal{I} \times \mathbb{R} \times \mathcal{E} \rightarrow \mathscr{A} $

\noindent
Having provided the definition of the generation operator, it is worth investigating its properties. Firstly, it should be noted that both internal and external constraints limit the number of the artefacts that the function can return. In particular, the image of the generation operator of a specific creative system is a proper subset of $\mathscr{A}$, and it is called the {\it potential generation space} $ \mathscr{A}_x^p$. This is meant to describe the artefacts that could potentially be returned by the generation operator, while considering $\mathcal{I}_x,\mathcal{E}_x$ as input. Intuitively, a creative system cannot intentionally generate something which goes beyond its external constraints and, of course, beyond its internal constraints. For instance, a human painter would not be able to produce a micron-sized paint-fleck; for similar reasons, an AI creative system is not able to generate artefacts whose size exceeds its memory capacity. Considering external constraints, a musician growning up in Africa and exposed only to African music cannot generate Western classical-like music unless she begins to listen to/study Western classical music. 

In order to provide a meaningful formal description, we rely on the notation introduced in \cite{coglione}, in which the symbol $\textcyrillic{I}$ is used to indicate the iteration of a function. We describe the space $ \mathscr{A}_x^p$ as the result of infinite repetitions of the generation function, with the very same input. 

$ \mathscr{A}_x^p = \textcyrillic{I}^\infty \lfloor \mathcal{I}_s,\mathcal{E}_x \rceil_\mathit{g} = \cup_{i=1}^\infty a_i$ $|$ $a_i \in \mathscr{A}$

\noindent
Having described the potential generation space of a single creative system, we are now interested in understanding the potential generation space of creative systems that share the same internal constraints. In other words, we want to describe the potential generation space of creative systems which have same ``cognitive" constraints, but different ``cultural" constraints. Formally, such potential space can be described as the union of all possible $\mathscr{A}^p_i$, which is still a proper subset of $\mathscr{A}$. This derives directly from the fact that the presence of internal constraints limits the potential artefacts space. $ \mathscr{A}^p = \cup_{i = 1}^\infty \mathscr{A}^p_i$

\subsection{Evaluation Operator}

A creative system without an evaluation function is solely a ``generative system", with no knowledge of the quality of its artefacts. The aim of the evaluation operator is to prevent the creative system from providing low-quality artefacts, thus evaluating every output generated by $\lfloor \cdot \rceil_\mathit{g}$, according to a specific metric. This takes into account both internal and external constraints, as well as the artefact $a$. 

$  \lfloor a, \mathcal{I}_s, \mathcal{E}_s, \mathcal{M} \rceil_\mathit{v} = ( c, r )$ $|$ $ a \in \mathscr{A}, c \in \mathcal{C}, r \in \mathbb{R}_{[0,1]} $

$ \lfloor \cdot \rceil_\mathit{v} :  \mathscr{A} \times \mathbb{R} \times \mathcal{I} \times \mathbb{R} \times \mathcal{E} \times \mathcal{M} \rightarrow \mathcal{C} \times \mathbb{R}_{[0,1]} $

\noindent
Society has a strong impact on the way in which creative systems evaluate artefacts. Likewise, internal constraints affect the ability of a system to evaluate artefacts. In particular, artefacts that are outside the cognitive capacities of a system, cannot be evaluated. For instance, a human cannot evaluate the aforementioned micron-sized painting. For modelling this behaviour, we designed the output of the evaluation function as a pair $(c,r)$, where $c$ is a qualitative value indicating if the overall evaluation is {\it positive} ($+$), {\it negative} ($-$) or {\it non-decidable} ($null$). $r$ is a real number, ranging between 0 and 1, representing the strength of the qualitative value. A value of 1 indicates a very strong feeling towards the qualitative measure, while a value close to 0 indicates an extremely weak feeling. 

As with the generation operator, it is interesting to describe the shape of the evaluation space for a given creative system. This can be identified as the union of all pairs $(c,r)$ returned by the operator, as follows.

$ \mathscr{V}_x = \lfloor \mathscr{A}, \mathcal{I}_s,\mathcal{E}_x, M \rceil_\mathit{v} = \cup_{i=1}^\infty (c,r)_i $

\noindent
Given the fact that the evaluation output can be classified according to three classes (positive, negative, non-decidable), it is possible also to define the spaces of positive and negative evaluations of a creative system, as well as the space of non-decidable artefacts.

$ \mathscr{V}_x^+ = \{ (c,r)$ $|$ $ c=\{+\}, r \in \mathbb{R}_{[0,1]} \}$

$ \mathscr{V}_x^- = \{ (c,r)$ $|$ $ c=\{-\}, r \in \mathbb{R}_{[0,1]} \}$

$ \mathscr{V}_x^n = \{ (c,r)$ $|$ $ c=\{null\}, r = 0 \}$

\noindent
We also define the complete space of the evaluations of a creative systems as the union of these three spaces.

$ \mathscr{V}_x = \mathscr{V}_x^+ \cup \mathscr{V}_x^- \cup \mathscr{V}_x^n$

\noindent
It is then straightforward to define the space of {\it evaluable} artefacts of a creative system, i.e. those that have $c \neq null$, as follows. 
$ \mathscr{V}_x^{dec} = \mathscr{V}_x^+ \cup \mathscr{V}_x^-$


\noindent
Furthermore, it can be observed that there is no intersection between the identified spaces of evaluation of a creative system. This directly comes from the way in which the evaluation function has been defined, and also from the fact that we believe the evaluation function to be deterministic, at least from the point of view of the qualitative classification. 

$ \mathscr{V}_x^+ \cap \mathscr{V}_x^- \cap \mathscr{V}_x^n = \emptyset$

\noindent
Considering a group of creative systems which share the same internal constraints it is interesting to analyse how they evaluate a given artefact $a$. To do that, we pass to the evaluation operator the artefact, the specific configuration of internal constraints, all possible configurations of external constraints for the given systems, and its memory set. The output is a 3-tuple of probability density functions. In particular, there are two probability density functions for the positive and negative classes, which indicate the likelihood of the weight $r$ to take on a given value. Since the null qualitative measure is a discrete variable that can be either true or false, we use a probability mass function, which returns the likelihood that the artefact is non-decidable.     

$ \lfloor a, \mathcal{I}_s,\mathcal{E}^T, M \rceil_\mathit{v} = ( \mathit{d}_\mathcal{C}(c), \mathit{f}_+(r), \mathit{f}_-(r)  ) $

\noindent
By incorporating in the above equation the entire artefacts space, rather than a single artefact, it is possible to derive the general space of evaluation for a class of systems that share the same internal constraints. The result is the union of infinite 3-tuples of probability density functions, which indicate the response of the systems to single artefacts.  

$ \mathscr{V} = \lfloor \mathscr{A}, \mathcal{I}_s,\mathcal{E}^T, M \rceil_\mathit{v} = \cup_{i=1}^\infty  ( \mathit{d}_\mathcal{C}(c), \mathit{f}_+(r), \mathit{f}_-(r)  ) $

\subsection{Update Operator}

Although it is not a necessary component of a creative system, the update operator allows the system to show T-creativity. In the proposed framework, the update operator accepts as input an artefact $a$, and the component of the creative system to update. It should be noted that $a$ can be generated by the same creative system, or it can be the product of other systems. The output is an updated version of the internal components. In particular, the generation operator, the evaluation operator and the configuration of external constraints $\mathcal{E}_s$ can be updated. Formally, the update operator is defined as follows:

$ \lfloor a, \lfloor \cdot \rceil_\mathit{v} \rceil_\mathit{u} = \lfloor \cdot \rceil_\mathit{v'}  $

$ \lfloor a, \lfloor \cdot \rceil_\mathit{v},  \mathcal{E}_s \rceil_\mathit{u} = \mathcal{E}_{s'}  $

$ \lfloor a, \lfloor \cdot \rceil_\mathit{v}, \lfloor \cdot \rceil_\mathit{g} \rceil_\mathit{u} = \lfloor \cdot \rceil_\mathit{g'} $

\noindent
In our formalisation, we consider internal constraints to be unmodifiable. Such constraints represent the cognitive limit of the creative system, that can hardly be changed. Although we acknowledge that this can happen in principle --training affects internal constraints, for instance--, modifications are extremely limited; and ignoring them does not invalidate the generality of the proposed framework.

\subsection{Special Cases}
As Wiggins suggests \cite{wiggins2006preliminary}, a theoretical framework not only proposes philosophical foundations, but also it is useful to help predict and describe general behaviours of creative agents. Specifically, we now discuss limiting cases, obtained by stretching the notion of creative system to its limit.

\hfill \linebreak
\noindent
{\bf Misunderstood genius}

\noindent
This creative system has no external constraints, i.e., $\mathcal{E}_s := \{ (0; e)$ $|$ $e \in \mathcal{E} \}$, and its creative capacity is limited only by internal constraints. Given enough time, the misunderstood genius is able to generate all possible artefacts, within a specific configuration of internal constraints. Having no external constraints is equal to having all possible configurations of external constraints $\mathcal{E}^T$. At first sight, the lack of external constraints seems a great advantage. However, the work of the misunderstood genius is hardly recognised by its peers, because their external constraints, which are usually quite similar within a social group, do not allow them to deem its artefacts as creative. A mild version of the misunderstood genius are those human innovators, like Van Gogh and late Beethoven, whose external constraints were by far different from those of their contemporaries. They found it difficult having their unconventional artefacts recognised as valuable by society, and perhaps, many of them are now forgotten, because of of their unusual creative process.

\hfill \linebreak
\noindent
{\bf Always (un)satisfied}

\noindent
These systems judge either as always absolutely positive or as always absolutely negative the artefacts they evaluate. 

$ \mathscr{V}_x = \lfloor \mathscr{A}, \mathcal{I}_s,\mathcal{E}_x, M \rceil_\mathit{v} = \cup_{i=1}^\infty (positive,1)_i $

$ \mathscr{V}_x = \lfloor \mathscr{A}, \mathcal{I}_s,\mathcal{E}_x, M \rceil_\mathit{v} = \cup_{i=1}^\infty (negative,1)_i $

\noindent
This behaviour completely inhibits the creative process of a system, since claiming that all artefacts are perfect equates to having no evaluation at all. In other words, an always (un)satisfied system is comparable to a ``generative system", which outputs artefacts randomly selected from its potential generation space. 

\hfill \linebreak
\noindent
{\bf Finite generator}

\noindent
A finite generator has a potential generation space which is finite: $\mathscr{A}^p = \{a_1,a_2,\dots,a_n\}$. Given enough time, the system eventually produces all of its possible creative outputs. In this case, the system appears to be creative until it generates all possible artefacts. After that moment, a finite generator ceases to be creative. Of course, the cardinality of the potential generation space depends on the strength and number of internal constraints. If these are too numerous and limit the system too much, it might be that the available artefacts are just a few. An interesting philosophical question is to consider whether or not the potential generation space of humans is finite. However, from a practical standpoint the space appears infinite to humans, since its cardinality is so big that it is impossible for a single person to generate all the artefacts of her space.

\hfill \linebreak
\noindent
{\bf Random walk}

\noindent
These 
%
%
systems update their internal state without taking actual artefacts as input. The updating process is completely random, and the result is a non-oriented wandering of the shape of the evaluation and the generation operators, as well as of the external constraints. Indeed, the key to updating a system is by considering the actual artefacts that have been previously evaluated. The updating process is similar to a learning process, which needs a continuous flow of raw materials in order to experiment with new things by trial and error. Random walk systems cannot learn over time, and are only a more hidden form of ``generative systems" with a small stress on evaluation, which in the end becomes completely random.



\section{From Systems to Society}

The generation of artefacts within the human domain depends greatly on relations between people. A person is influenced by the creative output of other people, and at the same time influences the creative behaviour of the people who happen to look at her artefacts. 
A society of creative systems is the natural expansion of a single system. In the next sections, we discuss how these societies might look like and we outline the rules that govern them.  

\subsection{Networks of Creative Systems}

Complex networks are graphs with non-trivial topology, which are used to model real-world systems such as organisations, the Internet and social systems \cite{nets}. The relations occurring between creative systems are based on the mutual exchange of artefacts in a non-linear, usually local manner, among several systems. This behaviour can be easily modelled using a complex network. 
Our framework describes society by exploiting {\it scale-free} networks \cite{barabasi}. This is the case, since most complex systems based on exchange of information in the real world, such as the World Wide Web, the human brain and market investment networks, can all be modelled with this type of graph \cite{caldarelli}. Also, some  properties of scale-free networks are particularly relevant for modelling a society based on evaluation of artefacts, and consecutive update of the states of its components. 

Specifically, scale-free networks are characterised by {\it preferential attachment}, which entails the {\it rich get richer effect}, where highly connected nodes increase the number of connections at the expense of other less connected components \cite{barabasi}. This phenomenon has been extensively observed within the human domain, and it is intrinsically guaranteed in a society in which information exchanged is assessed, and connections with far nodes are easy to make \cite{barabasi}. In the creative society, few systems are hyper-connected hubs with a large amount of connections, which drive the style of generation of the other, less influential agents. Nodes with fewer connections that oppose the creative status quo are usually marginalised, and seldom influence the entire network, by significantly changing the generation/evaluation properties of society. This argument is valid for any type of creative process, regardless of the differences between different domains, such as music, science or the visual arts.

In the near future, the creative society might contain both human and computational systems altogether. Indeed, it is probably not too far in the future when machines will be considered as creative agents, rather than tools \cite{futuro}. With this scenario, it is interesting to discuss a possible topology for the creative network. Given the difference between internal and external constraints of computers and humans, it might be the case that they will form two separated sub-graphs within the network; if machines are given the possibility to generate artefacts, which suit their constraints. An irreducible divergence will arise due to the difference between humans' and computers' internal constraints. 

\subsection{Generation and Evaluation}
For the time being, it is possible to identify three categories of creative systems: humans, computers and computer-aided systems. Even if there are relevant internal differences between elements of the same category, still the structure of internal and external constraints is usually similar between systems of the same class. The elements of these three categories considered together form the creative social network described in the previous section. It is possible to identify the generation and the evaluation spaces for each of these homogeneous categories.

The generation space for humans is formally defined as $ \mathscr{A}_{Hum}^p = \cup_{i=1}^n \mathscr{A}_i^p $. where $\mathscr{A}_i^p$ is the potential generation set of human creative systems considered as a whole. The space is the set union of the potential generation spaces of all $n$ humans inhabiting Earth at a certain time. $ \mathscr{A}_{Hum}^p $ depends on $ \mathcal{I}^{Hum}, \mathcal{E}^{Hum}$, which are the sets of all possible configurations of internal and external human constraints. Likewise, for computational creative systems we hold that $ \mathscr{A}_{Ccs}^p = \cup_{i=1}^n \mathscr{A}_i^p $ identifies the potential generation set of computers. $ \mathscr{A}_{Ccs}^p $ depends on $ \mathcal{I}^{Ccs}, \mathcal{E}^{Ccs}$, which are the sets of all possible configurations of internal as well as external computational constraints. Computational systems can visit regions of the potential generation space faster than humans, since they are potentially able to modify their external constraints quicker than humans \cite{futuro}. In this sense, computational creative systems are more flexible than humans. For computer-aided systems, the potential generation space is described by the equation $ \mathscr{A}_{Cad}^p = \mathscr{A}_{Hum}^p \cup \mathscr{A}_{Ext}^p$, where $\mathscr{A}_{Ext}^p$ is the set of artefacts that humans are not able to generate without computer-aided systems. The following is always true:

$\mathscr{A}_{Ext}^p \subset \mathscr{A}$, $\mathscr{A}_{Ext}^p \cap \mathscr{A}_{Hum}^p = \emptyset$

\noindent
The set union of potential generation spaces of all three categories are a subset of the potential generation space $\mathscr{A}$, since all systems have internal constraints that reduce the number of artefacts that can be generated.   

$ ( \mathscr{A}_{Hum}^p \cup  \mathscr{A}_{Cad}^p \cup  \mathscr{A}_{Ccs}^p) \subset  \mathscr{A}$

\noindent
The evaluation space for all humans is formally defined as:

$\mathscr{V}_{Hum}= \lfloor \mathscr{A}, \mathcal{I}^{Hum}, \mathcal{E}^{Hum} \rceil_\mathit{v} = \cup_{i=1}^\infty ( \mathit{d}_\mathcal{C}^{Hum}(c), \mathit{f}_+^{Hum}(r), \mathit{f}_-^{Hum}(r)  )$ 

\noindent
in which the probability density functions are calculated over all humans. A single 3-tuple represents the evaluation for a single point of the space.
A similar equation is used to calculate the evaluation space of computers:  

$\mathscr{V}_{Ccs}= \lfloor \mathscr{A}, \mathcal{I}^{Ccs}, \mathcal{E}^{Ccs} \rceil_\mathit{v} = \cup_{i=1}^\infty ( \mathit{d}_\mathcal{C}^{Ccs}(c), \mathit{f}_+^{Ccs}(r), \mathit{f}_-^{Ccs}(r)  )$

\section{General Creativity}
There is not a unique form of creativity; rather, there exist several forms of creativity, according to the types of creative systems involved in the generation and evaluation processes. In particular, a form of creativity is defined by a pair $\langle  \lfloor \cdot \rceil_\mathit{g},  \lfloor \cdot \rceil_\mathit{v} \rangle$ of generation and evaluation operators. 
Based on the evaluation operator, it is possible to identify two forms of creativity: anthropocentric and non-anthropocentric. If we consider these two forms together, we obtain General Creativity. In anthropocentric creativity the evaluation process is carried out by human evaluators, whereas for non-anthropocentric creativity to happen, non-human creative systems should be responsible for assessing artefacts. 

\subsection{Anthropocentric Creativity}
Anthropocentric creativity is the only form of creativity that has been extensively studied. Psychologists, philosophers and computer scientists have spent a great deal of time trying to unravel the mystery behind this type of creativity. Humans are central to anthropocentric creativity, since the final aim of it is to generate artefacts that can be recognised as novel and valuable from the human perspective. The only necessary constraint to have anthropocentric creativity is the evaluation operator. In other words, it is possible to use any conceivable generation strategy, as long as the result is appealing to a human evaluator. Thus, anthropocentric creativity can be divided into three different subcategories, according to the generation technique adopted: {\it humans for humans}, {\it computer-aided for humans}, {\it AI for humans}.

\hfill \linebreak
\noindent
{\bf Humans for humans (2H)} 

\noindent
In this form of creativity, $G$ is the 
human generation operator. This is a primary form of creativity, since it involves the creation of artefacts by humans, intended for other humans. Of course, before the advent of machines and computers this was the only form of creativity possible. Hence, 2H has been analysed for a long time, and it is the principal domain of study for psychologists interested in creativity. 

\hfill \linebreak
\noindent
{\bf Computer-aided for humans (CH)} 

\noindent
This form is characterised by a generation process based on the synergy between humans and machines. CH involves the generation of artefacts still valuable for humans that, are generated by humans with the support of a computational system. This type of creativity is extensively explored by performers and computer scientists interested in developing hybrid systems which exploit human-machine interaction to broaden the artefacts space available to humans. In this sense, computer-aided systems are usually physical extensions of performers which increase their potential generation spaces. 

\hfill \linebreak
\noindent
{\bf AI for humans (AIH)} 

\noindent
This form of creativity is the primary domain of study of computational creativity. It consists of artificial systems that are able to generate artefacts which might be deemed as creative by human evaluators. Of course, the generation process for AI anthropocentric creativity is not necessarily related to the way in which humans would create an artefact. Rather, the generation process exploits all possible techniques of AI and artificial life such as generative grammars, machine learning and evolutionary algorithms. 

\subsection{Non-anthropocentric Creativity}

Until now, the only form of creativity considered by researchers has been that in which the end users of artefacts are humans. However, by changing the type of creative system which evaluates an artefact it is possible to come up with a radical shift in the form of creativity. In particular, if we consider non-human creative systems as evaluators, it is possible to go far beyond anthropocentric creativity.

\hfill \linebreak
\noindent
{\bf AI for AI (2AI)} 

\noindent
This form of creativity is formally defined by the pair $\langle  \lfloor \cdot \rceil_\mathit{g}^{Ccs}, \lfloor \cdot \rceil_\mathit{v}^{Ccs} \rangle$, in which both components are operators 
of a computational system. This appears as a seismic change of perspective in the way creativity is conceived, which has not yet been explored  by researchers. 2AI focuses on artefacts which are generated by computational systems, meant to be assessed by other computational systems, with specific internal and external constraints. But, why should we care about a form of creativity we are not likely to understand at all? There are at least two good reasons to develop 2AI computational systems. First, by analysing complex creative systems beyond anthropocentric creativity we might discover a great deal of information about how anthropocentric creativity itself works. Second, this is a way to find and test the boundaries of the human evaluation operator, while discovering unpredictable creative results which could extend the human notion of creative artefacts. For these reasons, we encourage researchers interested in computational creativity to develop 2AI systems, so that they could go beyond anthropocentric creativity. In the long run, this new attitude might result in a significant shift in the general perception of creativity within other fields, such as psychology and philosophy. 







\section{Conclusions}

Computational creativity is a subfield of AI focused on building and studying the behaviour of creative systems. Most of the work done in the last decades by researches in the AI area aimed at providing agents able to generate creative artefacts. Very few studies introducing theoretical frameworks that describe the properties of creative agents on an abstract level have been undertaken until now. Also, such frameworks model anthropocentric creativity only. Since machines have completely different constraints from humans, they are potentially able to go beyond anthropocentric creativity, by exploring vast regions of the artefacts space unreachable to humans. Finally, they do not allow to clearly evaluate and simulate the interactions of societies of creative agents. 

In this paper we introduced General Creativity, which includes both anthropocentric and non-anthropocentric creativity. From this notion we developed a theoretical framework which can be used to describe creative agents and their interactions within a society. The framework characterises any type of creative system and can be easily extended in the future, if new types of creative agents arise.
By exploiting this framework we have been able to identify four different forms of creativity: humans for humans, computer-aided for humans, AI for humans, and AI for AI. 

Up until now, a great effort has been expended researching the first three instances of creativity, i.e. anthropocentric creativity. AI for AI has never been studied. By unleashing the power of AI creative systems we might develop a better understanding of how anthropocentric creativity actually works. Furthermore, we could discover the limits of human evaluation capabilities.

We see several avenues for future work. We plan to to deeply investigate specific computational creativity areas, such as music composition and narrative generation. This can be done by exploiting the proposed framework for modelling different existing AI-based creative approaches, and testing how do they interact and affect each other. Also, we plan to exploit the proposed framework for simulating the human society, and its behaviour with regards to creativity in the different areas of arts.

\bibliographystyle{plain}
\bibliography{ijcai15}

\end{document}